\documentclass[runningheads]{llncs}
\usepackage[T1]{fontenc}
\usepackage{graphicx}
\usepackage{orcidlink} 
\usepackage{bbding} 
\begin{document}
\title{Value Elicitation for a Socially Assistive Robot Addressing Social Anxiety: A Participatory Design Approach}
\titlerunning{Value Elicitation for a Socially Assistive Robot Addressing Social Anxiety}
%
\author{Vesna Poprcova$^{(}$\Envelope$^{)}$ \orcidlink{0009-0001-1504-6488} \and
Iulia Lefter \orcidlink{0000-0002-7243-2027} \and
Martijn Warnier \orcidlink{0000-0002-4682-6882} \and
Frances Brazier \orcidlink{0000-0002-7827-2351}}
\authorrunning{V. Poprcova et al.}
%
\institute{Delft University of Technology, Delft, The Netherlands \\
\email{\{v.p.poprcova,i.lefter,m.e.warnier,f.m.brazier\}@tudelft.nl}}
%
\maketitle              
\begin{abstract}
Social anxiety is a prevalent mental health condition that can significantly impact overall well-being and quality of life. Despite its widespread effects, adequate support or treatment for social anxiety is often insufficient. Advances in technology, particularly in social robotics, offer promising opportunities to complement traditional mental health. As an initial step toward developing effective solutions, it is essential to understand the values that shape what is considered meaningful, acceptable, and helpful. In this study, a  participatory design workshop was conducted with mental health academic researchers to elicit the underlying values that should inform the design of socially assistive robots for social anxiety support. Through creative, reflective, and envisioning activities, participants explored scenarios and design possibilities, allowing for systematic elicitation of values, expectations, needs, and preferences related to robot-supported interventions. The findings reveal rich insights into design-relevant values—including adaptivity, acceptance, and efficacy—that are core to support for individuals with social anxiety. This study highlights the significance of a research-led approach to value elicitation, emphasising user-centred and context-aware design considerations in the development of socially assistive robots.

\end{abstract}
\section{Introduction}

Social anxiety (SA) is a common and often debilitating mental health condition, characterised by intense discomfort in social situations and heightened fear of being judged by others \cite{stein_social_2008}. It can interfere with daily functioning, reduce quality of life, negatively affect overall well-being, and contribute to long-term emotional issues \cite{jefferson_social_2001}. Despite effective treatments, many who experience SA do not seek or receive adequate support due to stigma and limited access to care \cite{jefferson_social_2001,stein_social_2008}. 

Recent advances in Socially Assistive Robots (SARs) offer new opportunities to support mental healthcare \cite{riek_chapter_2016}. SARs are designed to interact through social behaviours and communication \cite{feil-seifer_defining_2005} and hold potential for addressing SA by creating a sense of being heard and understood, both crucial elements of emotional support \cite{laban_sharing_2024}. However, meaningful application in this area requires a deep understanding of what kinds of interactions and functionalities are desirable, helpful, and aligned with users' values.

As a starting point, it is essential to ground the design process in a human perspective to ensure that the technology aligns with human values, needs, and expectations. A participatory design approach is a methodology that promotes active participation of representative users throughout the design process \cite{muller_participatory_2002}. It often begins by collecting preliminary information through interviews or workshops to inform the design process before it starts, providing multiple benefits, e.g. to explore user needs and values, question existing assumptions, initiate knowledge exchange, and generate creative initial ideas \cite{fischer_symmetry_1999,muller_exploring_2017,muller_participatory_2002}. When users are involved as co-designers, designers/researchers improve their problem-solving capabilities and their potential to generate meaningful value \cite{fischer_symmetry_1999}.

This study presents the findings of a participatory design workshop that explore how SARs could be used to support those who experience SA, and the values that are core to their use. By focusing on users’ ideas and creativity, this study enabled the elicitation of core values, revealing design directions for future development of SARs.

\section{Background}

In this section, a review of SARs in the context of SA is presented, highlighting their potential. It also discusses the importance of integrating Value-sensitive design (VSD) to ensure that these technologies are effective in addressing SA, while remaining ethically grounded and responsive to users’ values. 

\subsection{\textbf{Socially Assistive Robots (SARs) for Social Anxiety (SA)}} 

SARs are advanced technological systems designed to provide support and promote positive outcomes by combining assistance with social interaction, aimed specifically to support users through non-physical means \cite{feil-seifer_defining_2005}. Such support often involves emotional engagement and interaction through speech and gestures \cite{feil-seifer_defining_2005}. Empirical evidence suggests that the advantages of self-disclosure to social robots extend beyond merely providing relief, by offering opportunities for emotional interventions that simulate understanding and provide feedback, thereby creating a sense of being heard and understood \cite{laban_sharing_2024}. 

Rasouli et al. \cite{rasouli_co-design_2025} co-designed a robotic mental well-being coach to help university students manage their public speaking anxiety. They conducted co-design sessions with mental health professionals to identify design-related needs and gathered feedback from university students to further refine the robotic coach. Their focus was on understanding users’ perceptions of a social robot as a coach for managing public speaking anxiety and their results revealed promising improvements in mood and relaxation during the session. Another study utilised a research-through-design approach to create Ommie, a social robot with haptic feedback designed to support deep breathing practices for anxiety reduction \cite{matheus_ommie_2025}. This study showed a reduction in anxiety levels and reported that interacting with Ommie was perceived as a highly engaging experience.  While the use of SARs for SA remains limited, existing research demonstrates their potential with vulnerable populations such as the elderly with dementia \cite{fardeau_impact_2023,ghafurian_social_2021} and children with autism \cite{santos_applications_2023}. In contrast to prior work centred on functional SAR design \cite{bradwell_design_2021,broadbent_acceptance_2009}, this study systematically elicits the underlying values for SA interventions through participatory design with experts.

\subsection{\textbf{Value-sensitive Design (VSD)}} 

Designing technologies for vulnerable populations, particularly those experiencing mental health conditions such as SA, requires careful attention to human experiences and values as well as ethical concerns. Value-sensitive design (VSD) is a theory-driven established approach to the design of technology that systematically integrates human values in a principled manner throughout the design process \cite{friedman_value_2019,umbrello_mapping_2021}. VSD is considered a part of the responsible research and innovation (RRI) approach, that has already been applied to varying extents to the design of SARs \cite{schmiedel_towards_2023,stahl_ethics_2016}. However, there is still potential for improvement to ensure outcomes that are both acceptable and desirable \cite{stahl_ethics_2016}. According to the AI for Social Good (AI4SG) VSD framework \cite{umbrello_mapping_2021}, SARs need to operate transparently and in an explainable manner, embedding critical values such as situational fairness and prevention of harm \cite{umbrello_value_2021}. Placing the human at the centre of design \cite{nevejan_presence_2014}, taking into account the interplay of values, interpersonal dynamics, and the socially situated nature of the complex system \cite{brazier_complex_2018}, have the potential to unlock new opportunities for research and design \cite{brazier_vision_2014,cheon_integrating_2016,ehsan_human-centered_2020}. 

Friedman and Kahn \cite{friedman_human_2002} have proposed a tripartite approach to VSD that consists of conceptual, empirical, and technical investigations, forming an iterative design process. A design framework for integrating values and ethics into the actual design process of social robots was developed by \cite{van_wynsberghe_designing_2013} using the conceptual and technical parts of the tripartite VSD methodology to integrate values into the design of care robots. The conceptual part elicits values through stakeholder-engaged investigations, using methods such as envisioning \cite{friedman_envisioning_2012} to explore values involved.

\section{Method}

To explore how SARs can be used to support people experiencing SA, a participatory design approach grounded in conceptual VSD was employed. This approach consisted of a workshop in which participants engaged in interactive activities to collaboratively envision the roles, and capabilities of a social robot designed to recognise and respond to SA. Impressions, ideas, and preferences were shared by participants, providing insights into how such technology could be meaningfully integrated into real-world contexts. The workshop aimed to generate actionable knowledge to elicit values to inform the design of SAR interventions tailored to enhance emotional well-being and address SA through a supportive and empathetic robotic companion.

\subsection{\textbf{Participants}}

Ethical approval for the workshop was granted by the Human Research Ethics Committee of TU Delft; participants took part voluntarily and provided informed consent. The workshop involved eight participants, primarily academic researchers with technical expertise in developing and studying technologies designed to advance mental healthcare. Self-selected participants formed two groups, each exploring the theme of SARs for SA support. One group consisted of two males and two females, and the other group consisted of four females. All participants held at least an MSc degree. 

\subsection{\textbf{Workshop}}

The workshop, lasting approximately 70 minutes, was designed to elicit design ideas for a socially assistive robot to support people experiencing social anxiety. A detailed description of the setting and activities is provided below. 

\begin{figure} [ht]
 	\centering 
 	\includegraphics[width=1\textwidth]
 {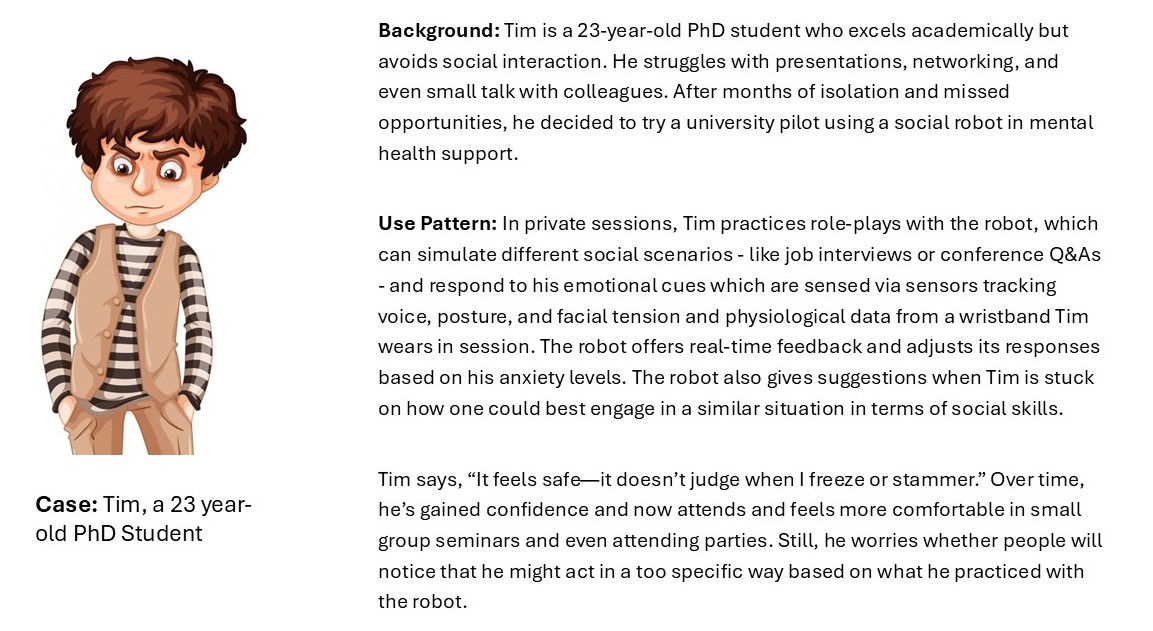}	
 	\caption{Background Story Used to Guide the Workshop} 
 	\label{fig1}
\end{figure}

\textbf{Scene setting (10 minutes).} After forming the groups, materials were distributed. Each group began to work on a detailed story as a starting point (see Fig.~\ref{fig1}) and a mind map using the vignette as a foundation. This vignette introduced Tim, a socially anxious PhD student who struggles with presentations and everyday interactions. To address this, he engages in role-play sessions with a socially assistive robot that uses advanced sensing technologies to simulate realistic social scenarios and provide real-time, adaptive feedback that supports gradual improvement. Subsequently, a slide was shown that displayed an image of a mind map, followed by three guiding questions for consideration about themes such as risks, benefits, and design ideas. Following \cite{mcdavid_eliciting_1985}, we employed an indirect way of elicitation to make participants feel more comfortable during the workshop and stimulate open discussions.  

The \textbf{goal} of the group-based activity was to encourage participants to explore the \textbf{risks} and \textbf{benefits} of using a socially assistive robot specifically designed for social anxiety, while envisioning \textbf{potential design solutions}. Subsequently, values were identified through analysis of participants' creative contributions to inform the design of SARs for SA support. 

\textbf{Activity 1: Ideation/Diverging (20 minutes).} Each group received a blank sheet of paper and was asked to individually write down their ideas using post-it notes. Afterwards, each participant received four marks to rate all concepts; they could distribute their marks freely without specific instructions or limitations to indicate which concepts in the mind map they deemed most important/inspiring for further research. 

\textbf{Activity 2: Discussion (20 minutes).} The two groups then discussed the rating scores and collectively selected the top-rated suggestions to present as their winning concept. 

\textbf{Activity 3: Presentation (20 minutes).} Each group created their own section of the idea gallery wall. To conclude, a representative from each group presented a summary of their group’s concept.

After the workshop, the completed sheets were collected for further analysis. 

\section{Analysis}

This section presents an analysis based on data collected during the participatory workshop. Further details regarding each group are provided in the following subsections.

\subsection{\textbf{Group 1}}

\begin{figure} [ht]
 	\centering 
 	\includegraphics[width=1\textwidth]
 {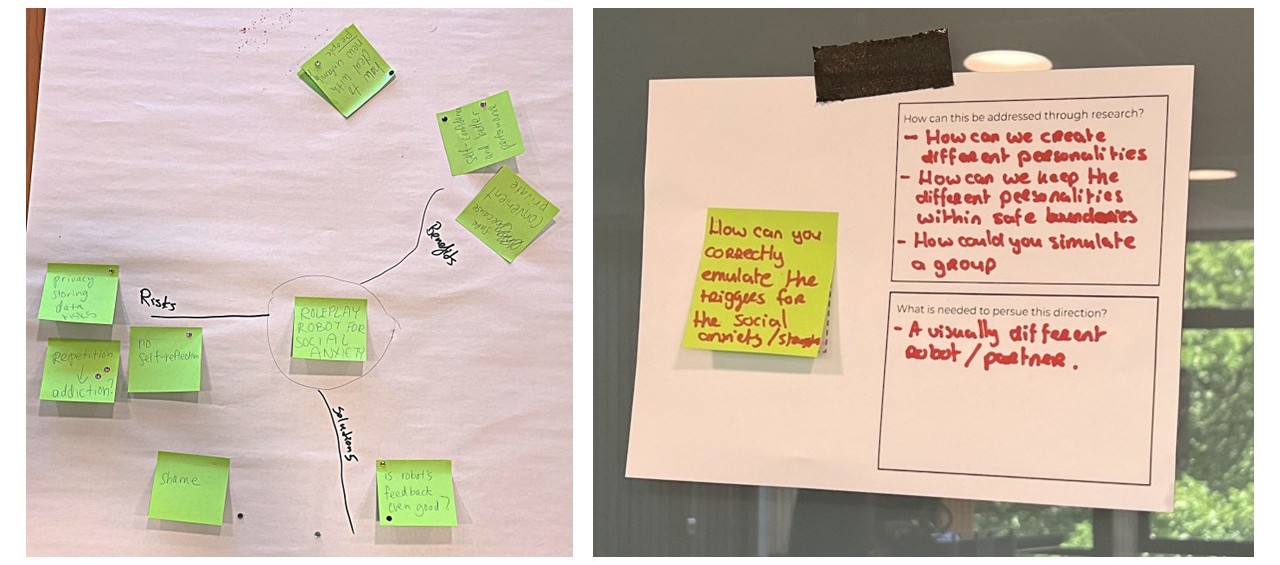}	
 	\caption{Group 1 Mind Map and Final Results} 
 	\label{fig2}
\end{figure}

An overview of first group's mind map (i.e.\ preliminary notes and ratings) is provided in Table~\ref{tab1} and Fig.~\ref{fig2}.

\begin{table}
\caption{Preliminary Details from Group 1}\label{tab1}
\centering
\begin{tabular}{|l|l|l|}
\hline
{\bfseries Risks} & Repetition; Addiction & High Importance \\
 & Privacy; Storing data risks & Medium Importance \\
 & No self-reflection & Medium Importance \\ 
 & Shame & Low Importance \\ 
\hline
{\bfseries Benefits } & Self-confidence and better performance & High Importance \\
 & How to deal with new unfamiliar people & High Importance \\
 & Different personalities, faces, and voices & Medium Importance \\
 & Safe because convenient and private & Low Importance \\
\hline
{\bfseries Design } & Is robot feedback even good? & High Importance \\
\hline
\end{tabular}
\end{table}

Group 1 (see Fig.~\ref{fig2}) focused on supporting the user in practising challenging or anxiety-inducing social situations. They formulated the central question: "How can you correctly emulate the triggers for social anxieties?". 

To address this, they proposed the following sub-questions:

\begin{itemize}
    \item How can we create different personalities?
    \item How can we keep the different personalities within safe boundaries?
    \item How can we simulate a group?
\end{itemize}

Participants highlighted the importance of creating different personalities for the robot, while ensuring these personalities remain within safe boundaries. They also emphasised the need to simulate group interactions, suggesting that recreating a social setting with multiple characters could better prepare users for real-world situations. Group 1 noted the need for "a visually different robot/partner" to pursue this research direction, suggesting that different types of visual cues could help enhance the realism of the interaction.

At the core of their concept is the aim to create a safe environment with which the user can face and work through anxiety-provoking social situations in a controlled and supportive setting. Over time, this could help the user develop greater confidence and more effective coping strategies. Importantly, their design considerations balance challenge and safety, reflecting an approach that integrates realism and user well-being.

\subsection{\textbf{Group 2}}

\begin{figure} [ht]
 	\centering 
 	\includegraphics[width=1\textwidth]
 {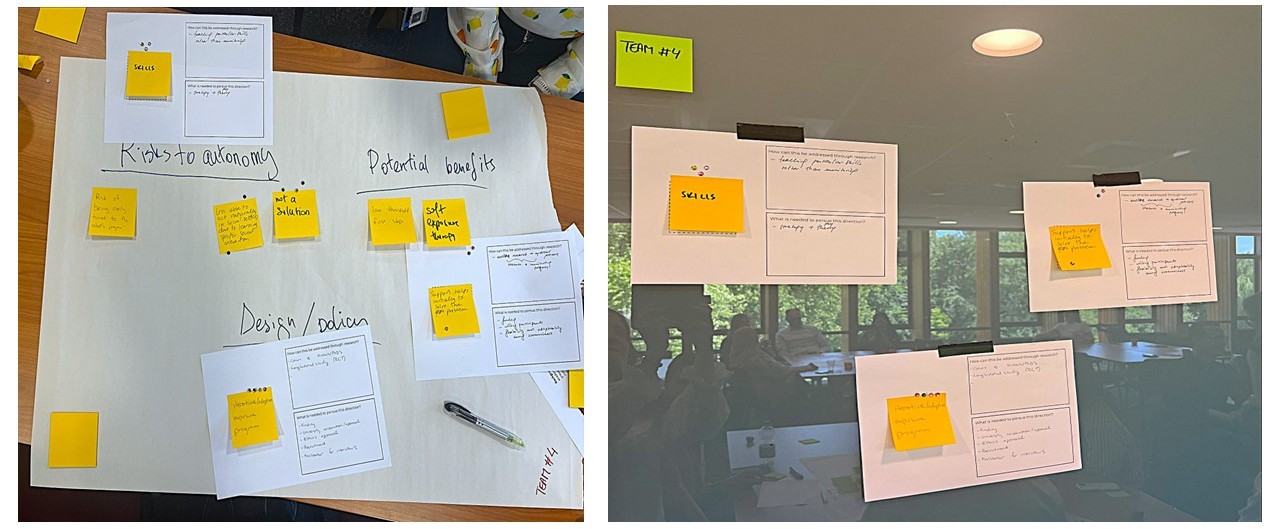}	
 	\caption{Group 2 Mind Map and Final Results} 
 	\label{fig3}
\end{figure}

An overview of Group 2's mind map (i.e.\ preliminary notes and rating) is provided in Table~\ref{tab2} and  Fig.~\ref{fig3}.

\begin{table}
\caption{Preliminary Details from Group 2}\label{tab2}
\centering
\begin{tabular}{|l|l|l|}
\hline
{\bfseries Risks} & Skills & High Importance \\
 & Not a solution & High Importance \\
 & Less able to act naturally in social settings & High Importance \\ 
 & Risk of being overly tuned to the robot & Low Importance \\ 
\hline
{\bfseries Benefits } & Support helps initially to solve the problem & High Importance \\
 & Low threshold first step & Low Importance \\
 & Self exposure therapy & Low Importance \\
\hline
{\bfseries Design} & Interactive/Adaptive exposure programme & High Importance \\
\hline
\end{tabular}
\end{table}

Group 2 chose to highlight and present the highest-rated concepts for each question category—risks, benefits, and design (see  Fig.~\ref{fig3}). 

Under the category "Risks" this team selected "skills" as an idea/requirement, which was ranked as most important. They proposed "teaching specific skills rather than monitoring" to address this risk through research, emphasising that what is needed is that the approach should be grounded in "psychology and theory". 

Considering the potential "Benefits" category, their selection of idea was "support helps initially to solve the problem". They stated addressing this through "action research and monitoring programme, in a cyclical process", identifying "funding, willing participants, and feasibility and adaptability among stakeholders" as essential prerequisites. 

For the final category, "Design", they proposed one requirement "iterative and adaptive exposure programme", which was considered of highest importance. Their suggestion for addressing this through research involved a "cohort of students/PhDs and a longitudinal study". According to this group, achieving this requires "funding, university cooperation and approval, ethical approval, participant recruitment, and a facilitator for interviews".

At the core of Group 2’s reasoning is the idea that SAR interventions should be research-driven, adaptive, and theory-informed, balancing risks and benefits to ensure meaningful skill-building and user empowerment. Their discussions illustrate how carefully considered design, grounded in psychological principles and supported by iterative evaluation, can inform actionable SAR design.

\section{Findings}

After completing the initial analysis, thematic analysis coding was conducted by the first author to identify key themes that represent values. The individual codes were then collaboratively discussed until agreement was reached on the most prominent findings supported by the data. This approach allowed in-depth examination of the 
gathered data, facilitating the identification of patterns and themes as proposed by \cite{braun_using_2006,vaismoradi_content_2013}.
Based on individual codes and subsequent code grouping, corresponding categories were identified similar to \cite{lyons_doing_2007,martynchamberlain_grounded_2013}.

\subsection{\textbf{Values Related to Perceived Risks}}

This subsection presents the findings in the form of the values elicited on perceived risks. The process identified patterns across the groups' inputs, supported by a ranking of perceived importance. Three themes emerged that highlight participants’ multi-dimensional concerns related to the expressed values that need to inform the design. Notably, the highest scoring risks focused on effectiveness and practical impact, suggesting that participants prioritised long-term human capability and authenticity in social interactions. This analysis provides critical insight into user-centred concerns that need to inform the design. 

\begin{figure} [ht]
 	\centering 
 	\includegraphics[width=1\textwidth]
 {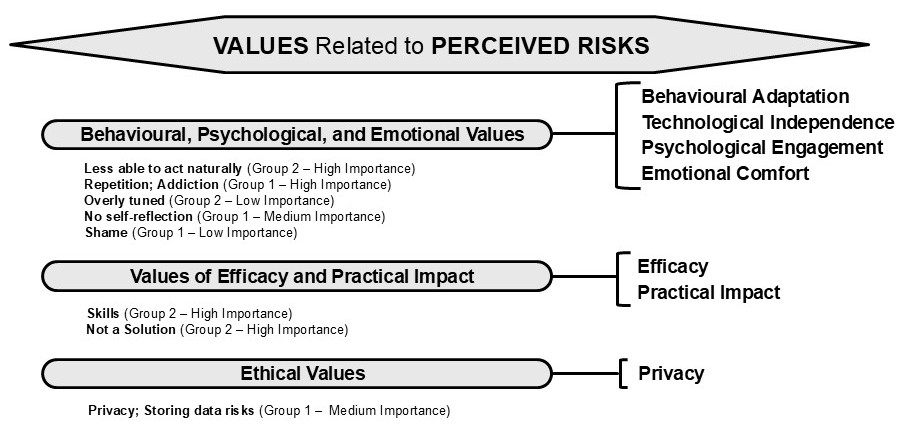}	
 	\caption{Connections between perceived risks and the elicited design values} 
 	\label{fig4}
\end{figure}

Fig.~\ref{fig4} illustrates the connections between perceived risks and the design values that were elicited.

\textbf{Behavioural, Psychological, and Emotional Values.} Participants expressed concern that users may become overly reliant, potentially developing problematic habitual behaviour (i.e.\ addiction) or becoming too accustomed to technological support. Overreliance on technology may lead to limited behavioural adaptation, restricting users’ ability to engage authentically in social interactions, while also limiting introspection and hindering personal growth. Emotional discomfort, such as shame, was mentioned as a potential risk; however, it was considered of lesser concern.

The following values were identified:

\begin{itemize}
     \item \textbf{Behavioural Adaptation} - "Less able to act naturally" (Group 2; High Importance);
     \item \textbf{Technological Independence} - "Repetition; Addiction" (Group 1; High Importance) and "Overly tuned" (Group 2; Low Importance);
     \item \textbf{Psychological Engagement} - "No self-reflection" (Group 1; Medium Importance); and 
     \item \textbf{Emotional Comfort} - "Shame" (Group 1; Low Importance).
\end{itemize}

\textbf{Values of Efficacy and Practical Impact.} Some participants perceived technological support to be surface-level or temporary. High rating scores suggest concern that such technologies might distract from more effective therapeutic interventions or delay the development of essential social skills. 

For this theme, the following values were identified:

\begin{itemize}
    \item \textbf{Efficacy} - "Skills" (Group 2; High Importance); and 
    \item \textbf{Practical Impact} - "Not a solution" (Group 2; High Importance).
\end{itemize}

\textbf{Ethical Values.} Ethical concerns about data privacy and surveillance were raised, including risks associated with data collection, storage, and the potential misuse of sensitive personal information.

This value was identified: 

\begin{itemize}
    \item \textbf{Privacy} - "Privacy; Storing data risks" (Group 1; Medium Importance).
\end{itemize}

\subsection{\textbf{Values Related to Potential Benefits}}

This subsection presents the themes that represent core areas from which participants' value elicitation emerged in relation to the benefits. These identified values emphasise the significance of building personal strength and confidence, developing useful skills, and feeling supported in a way that is both accessible and safe. They also highlight the importance of flexibility in responding to individual needs and changing circumstances. Together, these values reveal how benefits are most meaningful when they not only address immediate concerns but also contribute to a person’s overall sense of growth, security, and adaptability. 

\begin{figure} [ht]
 	\centering 
 	\includegraphics[width=1\textwidth]
 {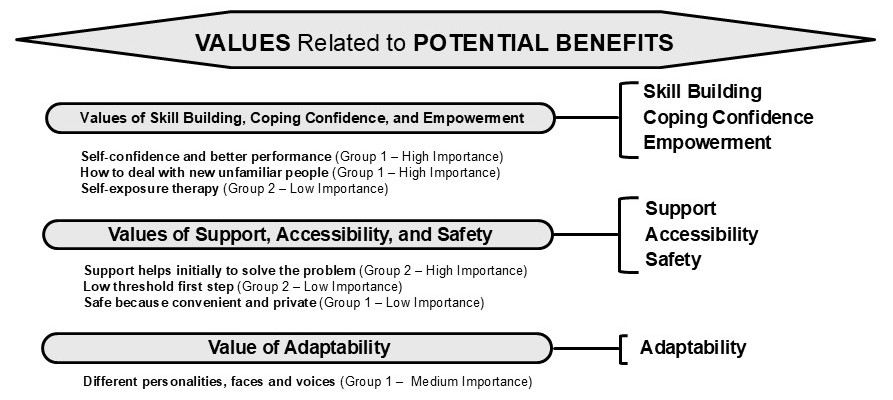}	
 	\caption{Connections between potential benefits and the elicited design values} 
 	\label{fig5}
\end{figure}

Fig.~\ref{fig5} illustrates the connections between potential benefits and the design values that were elicited. 

\textbf{Values of Skill Building, Coping Confidence, and Empowerment.} The participants emphasised that the robot could boost self-confidence and improve performance in social situations, offering the opportunity to practice and develop skills. In addition, the robot could help users develop coping strategies for interacting with new and unfamiliar people, serving as a form of self-exposure therapy to gradually reduce social anxiety.

The following values were identified:

\begin{itemize}
    \item \textbf{Skill Building}: "Self-confidence and better performance" (Group 1; High Importance); 
    \item \textbf{Coping Confidence}: "How to deal with new unfamiliar people" (Group 1; High Importance); and 
    \item \textbf{Empowerment} - "Self-exposure therapy" (Group 2; Low Importance).
\end{itemize}

\textbf{Values of Support, Accessibility, and Safety.} Participants highlighted that the robot could provide valuable initial problem-solving support, helping users address social anxiety in its early stages. They valued the robot’s ability to lower barriers to social support, describing it as a low-threshold, accessible first step for users hesitant to seek help. The safe, private, and convenient nature of the interaction was seen as essential, creating a secure environment that could encourage engagement without fear of judgment. 

The following values were identified:

\begin{itemize}
    \item \textbf{Support}: "Support helps initially to solve the problem" (Group 2; High Importance);
    \item \textbf{Accessibility}: "Low threshold first step" (Group 2; Low Importance); and
    \item \textbf{Safety}: "Safe because convenient and private" (Group 1; Low Importance).
\end{itemize}

\textbf{Value of Adaptability.} Participants noted that the robot’s ability to simulate different gestures, faces, and voices could enrich interactions, and potentially make them feel more realistic and personalised. This ability was seen as key to supporting users in practicing social skills across a range of contexts and social styles, which could better prepare them for real-world social encounters.

This value was identified: 

\begin{itemize}
    \item \textbf{Adaptability} - "Different personalities, faces and voices" (Group 1; Medium Importance). 
\end{itemize}

\subsection{\textbf{Values Related to Design}}

\begin{figure} [ht]
 	\centering 
 	\includegraphics[width=1\textwidth]
 {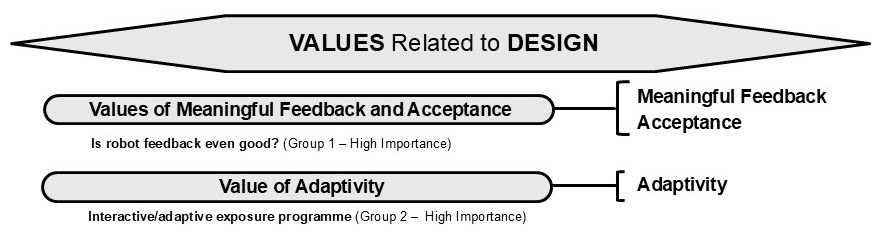}	
 	\caption{Connections between design ideas and the elicited values} 
 	\label{fig6}
\end{figure}

This subsection presents the themes in the form of the values elicited related to the design ideas. The analysis of the design ideas revealed the values of meaningful feedback, acceptance, and adaptivity. Fig.~\ref{fig6} illustrates the connections between design ideas and values that were elicited.

\textbf{Values of Meaningful Feedback and Acceptance.} Participants raised important design considerations that focused on the quality and acceptance of the robot, questioning whether it provides meaningful feedback and helpful responses. 

These values were identified:

\begin{itemize}
    \item \textbf{Meaningful Feedback} - "Is robot feedback even good?" (Group 1; High Importance); and
    \item \textbf{Acceptance} - "Is robot feedback even good?" (Group 1; High Importance).
\end{itemize} 

\textbf{Value of Adaptivity.} There was also strong support for incorporating interactive and adaptive exposure programmes that could tailor experience to individual needs, enhancing engagement and therapeutic benefit through personalised, dynamic interactions. 

This value emerged from the theme that has a highest-scoring coding:

\begin{itemize}
    \item \textbf{Adaptivity} - "Interactive/adaptive exposure programme" (Group 2; High Importance).
\end{itemize} 

\section{Discussion}

The findings reveal an interconnection between participant concerns, perceived benefits, and proposed solutions, each reflecting specific underlying values that shape expectations around the use of SARs for emotional well-being and support in SA contexts.

The identified risks underscore the core value of emotional comfort. Participants highlight the importance of designing systems that are both emotionally attuned and non-intrusive, and that enable technological independence. Similarly, values of efficacy, practical impact, and privacy emphasise the necessity of long-term impact evaluation and ethical oversight. The questions of which values are emphasised is relevant to any technology that interacts with humans \cite{ostrowski_ethics_2022}, and particularly critical for SARs, where marginalised values such as privacy \cite{brandao_normative_2021,warnier_design_2015} can easily be overlooked.

Themes related to benefits elicit values toward empowerment and inclusivity. Participants envisioned robotic companions not as replacements for human interaction, but as facilitators that could offer safe, low-pressure supportive and accesible environment to practice social engagement. The emphasis on adaptability further supports the value of authenticity, with participants highlighting the importance of tailoring interactions to individual needs and social contexts to enhance user trust and engagement. The co-designers in this study were mental health academic researchers, aware of the challenges involved.  The next step is to explore the values end users with SA embrace in participatory design sessions in different settings. This is particularly critical when the robot interacts directly with the user, as understanding the users' values is essential for providing appropriate support \cite{cranefield_no_2017}. An important aspect of this understanding involves emotions, given that users' affective responses to design and technology often reflect their underlying personal and moral values, thereby serving as a gateway to values \cite{desmet_emotions_2014}. Furthermore, conducting longitudinal co-design would allow assessment of sustained value alignment over time, while cross-cultural investigations \cite{lim_social_2021} could validate and enrich the set of elicited values.

The proposed designs centre on values of user-centredness, meaningful feedback, and co-creation. Participants stressed that meaningful engagement requires systems that adapt over time, reflect user feedback, and evolve, which aligns with participatory design values. Notably, the focus on adaptivity also suggests a value placed on human interaction to better simulate authenticity. Previous work in SARs similarly emphasises co-creation, often using workshops or brainstorming sessions involving stakeholders in the design process \cite{huijnen_how_2017,moharana_robots_2019}. In accordance with \cite{pommeranz_elicitation_2012}, this study aims to incorporate human values into the responsible design process \cite{poprcova_affect-sensitive_2024}, with a particular emphasis on eliciting values relevant to the context of SARs for SA support that could empower users to manage their social anxiety more effectively \cite{poprcova_exploring_2024}.

\section{Conclusion} 

In conclusion, this study highlights the critical role of value elicitation in informing the design of SAR interventions for SA support. Analysing participants' creative insights, key values emerged that reflect a strong emphasis on interactive design considerations, including an adaptive robot to effectively emulate social triggers. In addition, simulation of group interactions further reflects the importance of representing real-world social contexts to better prepare users for social challenges. The findings advocate the need for a theoretically grounded, research-led approach that integrates action research cycles and longitudinal evaluation. While the study was limited in participant diversity and scale, it drew upon the perspectives of academic researchers in mental health bringing valuable domain expertise and depth. This expertise reinforces the relevance of the elicited values—such as adaptivity, meaningful feedback, acceptance, support, coping confidence, skill building, efficacy, and practical impact (i.e.\ all derived from ideas marked as highly important)—and provides a basis for developing tailored SAR interventions. 

Overall, this study underscores the importance of value elicitation in shaping the design of SAR interventions aimed at supporting individuals with SA. Analysis of participants' creative insights revealed key values that suggest several design guidelines: 
\begin{itemize}
    \item SARs should be adaptive, responding dynamically to users’ emotional states and social cues; 
    \item they should provide meaningful feedback that guides users toward improved coping and social skills; 
    \item interactions should simulate realistic social contexts, including group scenarios, to prepare users for social challenges; and 
    \item SARs should be designed to enhance user confidence, offering complementary support and nonjudgmental engagement that encourage sustained practice. 
\end{itemize}

In a future study, the inclusion of individuals with SA as end users will be essential to ensure that SAR interventions are informed not only by expert knowledge but also by the values, lived experiences, and priorities of those directly affected.

Taken together, the findings indicate that participants' requirements for supportive SARs are not simply technical or functional in nature—they are deeply value-driven. Designing for this domain requires a thoughtful alignment between technological capabilities and the human values that are intended to support the nuanced needs of those with SA.

\bibliographystyle{splncs04}

\end{document}